\documentclass[lettersize,journal]{IEEEtran}
\usepackage{amsmath,amsfonts}
\usepackage{algorithmic}
\usepackage{algorithm}
\usepackage{array}
\usepackage[caption=false,font=normalsize,labelfont=sf,textfont=sf]{subfig}
\usepackage{textcomp}
\usepackage{stfloats}
\usepackage{url}
\usepackage{verbatim}
\usepackage{graphicx}
\usepackage{cite}
\usepackage{booktabs}
\usepackage{multirow}
\usepackage{amssymb}
\usepackage{subcaption}

\begin{document}

\title{Mode-as-Sequence: Translating Multimodal Motion Prediction into Unified Sequential Mode Modeling}


\author{
Zikang Zhou$^{1}$ \quad Haibo Hu$^{1}$ \quad Xinhong Chen$^{1}$ Yifan Zhang$^{2}$ \\
\quad \textbf{Nan Guan$^{1}$} \quad \textbf{Yung-Hui Li}$^{3}$ \quad \textbf{Chun Jason Xue}$^{4}$ \quad \textbf{Jianping Wang$^{1}$}\\
$^{1}$City University of Hong Kong \quad $^{2}$City University of Hong Kong (Dongguan)\\
$^{3}$Hon Hai Research Institute \quad $^{4}$Mohamed bin Zayed University of Artificial Intelligence\\
\thanks{Corresponding author: Haibo Hu (haibohu2-c@my.cityu.edu.hk)}
}

\markboth{Journal of \LaTeX\ Class Files,~Vol.~14, No.~8, August~2021}%
{Shell \MakeLowercase{\textit{et al.}}: A Sample Article Using IEEEtran.cls for IEEE Journals}

\maketitle

\begin{abstract}
Multimodal motion forecasting is inherently under-supervised: each training scene provides only one realized future, yet multiple plausible futures exist. This sparse supervision often leads to mode collapse (redundant hypotheses and insufficient mode coverage) and unreliable confidence ranking when predicting a small set of trajectories. We propose \textbf{Mode-as-Sequence}, a unified decoding framework that translates an unordered mode set into an ordered mode sequence and explicitly models mode-to-mode dependency. Under this framework, we develop two complementary instantiations. \textbf{ModeSeq} performs recurrent mode decoding, where each mode is generated conditioned on the previously generated modes, encouraging diverse, non-redundant hypotheses with calibrated confidence ordering. To remove the mode-by-mode autoregressive bottleneck, we further propose \textbf{Parallel ModeSeq}, which preserves the same causal dependency using masked mode-to-mode self-attention while decoding all modes in a single forward pass, enabling efficient large-$K$ inference and scalable joint-scene prediction. To learn representative modes and calibrated confidence under sparse labels, we introduce \textbf{Early-Match-Take-All} (EMTA) and its joint-scene extension \textbf{MA-EMTA}, together with a lightweight ranking regularizer that reduces confidence inversions. Extensive experiments on large-scale benchmarks demonstrate consistent improvements in both ranking-oriented metrics and best-of-$K$ accuracy across datasets, horizons, and object types. In the Waymo Open Dataset challenges, ModeSeq achieves \textbf{1st place} in the \textbf{2024 LiDAR-free} motion prediction track, and Parallel ModeSeq achieves \textbf{1st place} in the \textbf{2025} Interaction Prediction Challenge, validating the effectiveness of Mode-as-Sequence for both accuracy and efficiency.
\end{abstract}


\section{Introduction}
Handling the intrinsic uncertainty of real-world traffic is a central hurdle for autonomous driving.
A major source of such uncertainty lies in multimodal agent behaviors: conditioned on the same past observation and map context, multiple plausible futures (e.g., yield vs.\ proceed, turn vs.\ go straight) may all be valid.
Accurately forecasting a small set of representative trajectories with meaningful confidence scores is therefore critical for safe interaction and downstream decision making.
This challenge has been actively studied on large-scale benchmarks such as the Waymo Open Motion Dataset~\cite{ettinger2021large} and Argoverse~2.0~\cite{wilson2021argoverse} (as well as nuScenes~\cite{caesar2020nuscenes} and INTERACTION~\cite{zhan2019interaction}), where predictors must jointly balance trajectory accuracy, mode diversity, and ranking quality.
\begin{figure}[t]
  \centering
  \includegraphics[width=1\linewidth]{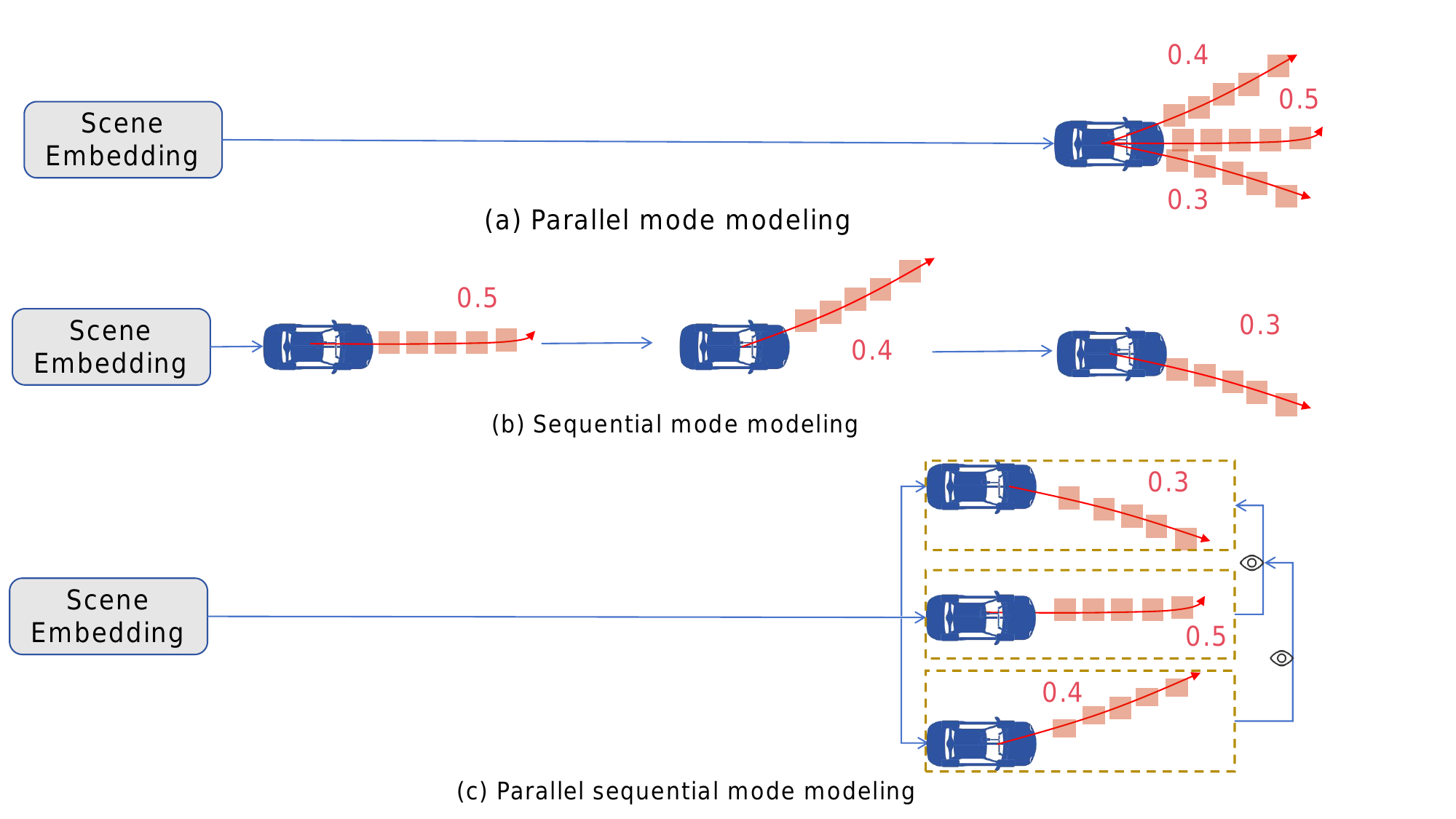}
\caption{Illustration of three different ways to model future trajectories: (a) Parallel Mode Modeling, where all future modes are generated simultaneously without inter-mode dependency; (b) Sequential Mode Modeling, where modes are generated autoregressively, with each mode conditioned on the previously generated ones; and (c) Parallel Sequential Mode Modeling, which combines parallel generation with causal inter-mode conditioning. In (c), the visual markers indicate the causal receptive field: the second mode can access the first mode, while the third mode can access both the first and second modes. The numbers around each trajectory indicate the mode confidence.}
    \label{fig:mode_comparison}
\end{figure}
Despite extensive progress in multimodal prediction~\cite{chai2019multipath,phan2020covernet,ivanovic2019trajectron,salzmann2020trajectron,gupta2018social},
learning a calibrated multimodal distribution remains fundamentally difficult because real-world data typically provides only one ground-truth future per scene.
A prevalent remedy is winner-take-all (WTA) or multiple-choice learning, where only the best-matched mode receives strong supervision while the remaining modes are weakly trained or ignored.
In practice, this often leads to (i) mode collapse (near-duplicate trajectories) and (ii) ambiguous confidence learning (indistinguishable scores among overlapped modes), as also discussed in prior multimodal modeling analyses~\cite{makansi2019overcoming,bhattacharyya2018accurate}.
To compensate for the missing multimodal supervision, many pipelines shift to generating a large pool of candidates (dense goals or excessive anchors) and then heuristically selecting a sparse representative subset, e.g., via rule-based post-processing such as non-maximum suppression (NMS)~\cite{zhao2021densetnt,gu2021densetnt,phan2020covernet}.
While dense prediction can improve coverage, it introduces two practical limitations.
First, post-processing requires careful hyper-parameter tuning that may not transfer across diverse scenarios.
Second, accurately extracting the truly representative modes from a large candidate set is itself non-trivial, and the selection step may sacrifice trajectory quality in realistic deployment.
These issues become even more pronounced for modern Transformer/graph predictors~\cite{zhou2022hivt,shi2023mtr,nayakanti2023wayformer,zhou2023query}, where the decoding complexity and ranking ambiguity grow with the candidate pool size.

Motivated by these limitations, we seek an end-to-end solution that directly outputs a sparse set of diverse and high-quality trajectories without dense candidates or heuristic selection.
We observe a shared property of mainstream multimodal decoders: they generate all $K$ modes in one shot (in parallel), treating each mode prediction largely independently.
In anchor-free designs, the distinction between modes often relies on head-specific parameters; under WTA-like training, this distinction is unstable and difficult to control, which encourages duplicated hypotheses.
In anchor-based designs, diversity is partially offloaded to anchors, but selecting a sparse anchor set that adapts to each scene is difficult---hence the tendency to use many anchors and revert back to dense prediction~\cite{chai2019multipath,phan2020covernet,zhao2021densetnt,gu2021densetnt}.
Overall, under the parallel mode modeling paradigm, producing diverse multimodal forecasts with sparse mode prediction faces intrinsic obstacles.
To address sparse multimodality at its root, we advocate a different decoding paradigm: sequential mode modeling (Mode-as-Sequence).
Instead of predicting an unordered set of modes in parallel, we construct a mode sequence and decode one mode at a time, where each new mode is conditioned on previously decoded modes.
This design explicitly captures mode-to-mode correlations: the model can reason about what futures have already been covered and allocate later modes to complementary behaviors, thereby improving diversity and reducing duplicates without relying on dense candidates, anchors, or post-processing.
Crucially, sequential mode modeling also induces a natural ordering among modes, which can be exploited to learn better confidence ranking.
In Figure~\ref{fig:mode_comparison}, we illustrate the difference between three mode prediction strategies: (a) Parallel Mode Modeling, where all future modes are generated simultaneously without dependency between them; (b) Sequential Mode Modeling (ModeSeq), where modes are generated sequentially with each mode conditioned on the previous one; and (c) Parallel Sequential Mode Modeling (Parallel ModeSeq), which combines the advantages of parallel generation with sequential mode conditioning, preserving causal dependencies between modes while ensuring efficient parallel inference.

Built upon the sequential formulation, we introduce an Early-Match-Take-All (EMTA) training strategy.
Intuitively, EMTA encourages the decoder to produce a mode that matches the ground-truth as early as possible in the mode sequence with high confidence.
Meanwhile, once a matched mode is produced, EMTA pushes the remaining modes to vacate duplicated futures and explore other plausible behaviors, improving mode coverage with negligible degradation in trajectory accuracy.
This simultaneously alleviates mode collapse and makes confidence learning easier, since earlier modes are explicitly trained to be more representative.

To further enhance modeling capacity under the new paradigm, we adopt an iterative refinement decoder that stacks multiple decoding layers.
Between layers, we perform mode rearrangement to align the mode order with the learned confidence ranking, coordinating the refinement process with EMTA and stabilizing the mode sequence structure.
This refinement-and-reorder strategy is compatible with query-based decoders~\cite{zhou2023query} and scales effectively to stronger encoders.
While recurrent sequential decoding provides an elegant mechanism for mode dependency modeling, its step-by-step generation can become a latency bottleneck when a large number of modes is required, and it is less convenient for joint multi-agent prediction.
To bridge this gap, we extend ModeSeq into Parallel ModeSeq: we preserve the same causal mode dependency but realize it with masked mode-to-mode attention, translating an unordered mode set into an ordered sequence and decoding all modes in one shot.
This parallel realization retains the key benefit of sequential mode modeling---explicit mode-to-mode coordination---while enabling GPU-friendly high-throughput inference.

Moreover, Parallel ModeSeq naturally supports scene-level multimodal prediction for interactions, avoiding the combinatorial explosion encountered by anchor-based joint modeling and mitigating the inconsistency issues of heuristic marginal-to-joint composition.
To further improve scoring quality, we incorporate a margin ranking objective to suppress confidence inversions, which is particularly beneficial for mAP-style evaluation.
We also generalize EMTA to multi-agent supervision (MA-EMTA), enabling end-to-end learning of diverse, representative, and well-ranked joint futures.

In summary, this paper makes the following contributions:
\begin{itemize}
  \item We introduce Mode-as-Sequence, a unified view that models multimodal futures via sequential mode dependencies, addressing sparse multimodality without dense candidates or heuristic selection.
  \item We develop ModeSeq with EMTA and a refinement-and-reorder decoder, improving diversity and confidence ranking while maintaining trajectory accuracy.
  \item We propose Parallel ModeSeq, a parallel realization of sequential mode modeling via causal mode-to-mode attention, enabling efficient inference and a principled extension to joint multi-agent prediction with MA-EMTA and margin ranking.
\item We validate the proposed framework on large-scale benchmarks~\cite{ettinger2021large} and obtain first-place results on the Waymo Open Motion Dataset challenges: ModeSeq wins 1st place in the 2024 LiDAR-free motion prediction track, and Parallel ModeSeq wins 1st place in the 2025 Waymo interaction prediction challenge.
\end{itemize}
Compared with our CVPR conference version on ModeSeq~\cite{zhou2025modeseq}, this submission substantially extends the method and analysis. The new contributions include:
1) a unified \emph{Mode-as-Sequence} formulation that systematically translates an unordered mode set into an ordered mode sequence and clarifies the common principles behind sequential/parallel decoding;
2) Parallel ModeSeq, an efficient parallel realization that preserves causal mode-to-mode dependency via masked attention, enabling scalable inference with large $K$ and supporting scene-level (joint) forecasting;
3) extended training objectives and evaluations, including MA-EMTA for joint-scene supervision and an additional ranking regularizer to reduce confidence inversions, together with more comprehensive experiments/ablations and challenge-level validations.

\section{Related Work}

\subsection{Multimodal Motion Prediction}
The evolution of motion prediction has been driven by the need to capture the inherent multimodality of driving behaviors while maintaining representation efficiency. Early approaches adopted rasterized representations, projecting the scene into Bird's-Eye-View (BEV) images processed by Convolutional Neural Networks (CNNs)~\cite{cui2019multimodal, chai2019multipath, djuric2020uncertainty, hong2019rules, bansal2018chauffeurnet}. To handle multimodality, these methods often employed anchor-based regression~\cite{chai2019multipath, phan2020covernet} or heatmap estimation~\cite{gilles2021home, gilles2022gohome}. However, rasterization suffers from quantization errors and high computational costs.

The field subsequently shifted towards vectorized representations, pioneered by VectorNet~\cite{gao2020vectornet}, which models map elements and agents as sparse polylines. Following this, LaneGCN~\cite{liang2020learning} introduced lane-graph convolutions to capture map topology. Building on these encoders, decoding paradigms have diverged into two main streams: goal-based and query-based. Goal-based methods, such as TNT~\cite{zhao2020tnt} and DenseTNT~\cite{zhao2021densetnt}, formulate prediction as goal candidate classification followed by trajectory completion. While effective, they depend heavily on the quality of goal sampling.

Recent state-of-the-art methods predominantly adopt a \textit{Parallel Mode Modeling} paradigm via Transformers. Methods like HiVT~\cite{zhou2022hivt}, MTR~\cite{shi2023mtr}, Wayformer~\cite{nayakanti2023wayformer}, and others~\cite{zhou2023query, varadarajan2022multipath} utilize learnable queries or intent embeddings to decode multiple trajectories simultaneously. Although these parallel architectures achieve high efficiency and performance on benchmarks like Waymo~\cite{ettinger2021large} and Argoverse~\cite{wilson2021argoverse}, they typically regress the entire temporal horizon in a one-shot manner. This design implicitly assumes mode independence, potentially overlooking the temporal evolution of intent that our work seeks to address.

\subsection{Sequential Modeling in Prediction}
Sequential modeling explicitly captures the temporal dependencies of motion, treating the future state as a function of previous states. Classical approaches utilized Recurrent Neural Networks (RNNs) and LSTMs to roll out trajectories step-by-step~\cite{alahi2016social, salzmann2020trajectron, ivanovic2019trajectron, deo2018convolutional}. For instance, Social-LSTM~\cite{alahi2016social} introduced pooling layers to model sequential interactions, while Trajectron++~\cite{salzmann2020trajectron} incorporated CVAEs~\cite{lee2017desire} within recurrent units to generate diverse futures. Despite their ability to model causality, autoregressive RNNs often suffer from error accumulation over long horizons.

More recently, the concept of sequential refinement has been revitalized by generative diffusion models. Methods like MID~\cite{gu2022stochastic}, MotionDiffuser~\cite{jiang2023motiondiffuser}, Leapfrog~\cite{mao2023leapfrog}, and LED~\cite{zhong2023guided} formulate prediction as a sequential denoising process. These approaches achieve superior mode coverage and controllability by iteratively refining trajectories from Gaussian noise. However, the heavy computational burden of multi-step denoising renders them less suitable for real-time onboard applications compared to regression-based methods. Our proposed Sequential Mode Modeling draws inspiration from these sequential reasoning processes but integrates them into a parallelizable query-based framework, balancing inference speed with the capability to capture evolving branching intents.

\subsection{Joint Interaction and Planning-Centric Prediction}
Effective motion forecasting must account for the mutual influence between agents (joint prediction) and the downstream utility of the predicted trajectories (planning-centric). To model complex agent-agent interactions, Scene Transformer~\cite{ngiam2021scene} and AgentFormer~\cite{yuan2021agent} flatten the temporal and agent dimensions, applying global attention to capture symmetric relationships. Other works leverage graph-based interaction networks~\cite{li2021rain, jia2023hdgt} or game-theoretic formulations~\cite{huang2022gameformer} to model negotiation and yielding behaviors explicitly.

Furthermore, there is a growing trend towards aligning prediction objectives with planning tasks. End-to-end frameworks like UniAD~\cite{hu2023planning}, VAD~\cite{jiang2023vad}, and classic imitation learning approaches~\cite{bansal2018chauffeurnet, sadat2020perceive, rhinehart2019precog} optimize prediction and planning jointly. These methods emphasize the importance of scene-level consistency and safety over simple distance metrics. In this context, our approach contributes by generating trajectories with temporally consistent confidence scores, which acts as a reliable cost prior for downstream planners~\cite{schwarting2018planning}, reducing the risk of "flickering" plans caused by unstable mode probabilities.

\section{Unified Multimodal Prediction Framework: From ModeSeq to Parallel ModeSeq}
In this section, we introduce the unified framework for multimodal trajectory forecasting, which seamlessly transitions from a sequential mode decoding approach (ModeSeq) to an efficient parallelized version (Parallel ModeSeq). We begin by formally defining the problem, followed by a motivation for our approach. Then, we introduce the key components of our framework, including single-agent and multi-agent prediction, as well as the training objectives.

\subsection{Problem Formulation and Generic Architecture}
\label{sec:problem_def}

\paragraph{Problem Formulation} 
 Without loss of generality, we discuss the formulation of single-agent prediction, which can be extended to the multi-agent setting via batchifying across agents. Denote $S$ as the holistic input for motion prediction, which encompasses the vectorized map elements represented as $M$ polygonal instances and the history trajectories of $A$ traffic agents (e.g., vehicles, pedestrians, and bicycles) over the past $T_{obs}$ steps. Each map polygon consists of line segments outlining the topology and semantic attributes describing the traffic rules. Similarly, agent trajectories are composed of state sequences (e.g., position, velocity, heading, and bounding box sizes) and agent types.

The goal is to forecast $K$ plausible trajectory modes for each target agent. Each mode $k$ consists of a future trajectory $\hat{\mathbf{y}}_k = [\hat{y}_{k}^1, \dots, \hat{y}_{k}^{\hat{T}}] \in \mathbb{R}^{\hat{T} \times 2}$ over the prediction horizon $\hat{T}$, and an associated confidence score $\hat{\phi}_k \in [0, 1]$. Note that the $K$ confidence scores are not strictly required to sum to 1, as their relative ranking is sufficient for downstream applications such as ego motion planning. The predicted modes should be representative, covering distinct maneuvers while accurately reflecting their likelihoods.
\begin{figure*}[t]
  \centering
  \includegraphics[trim=0 330 0 20, clip,width=1\textwidth]{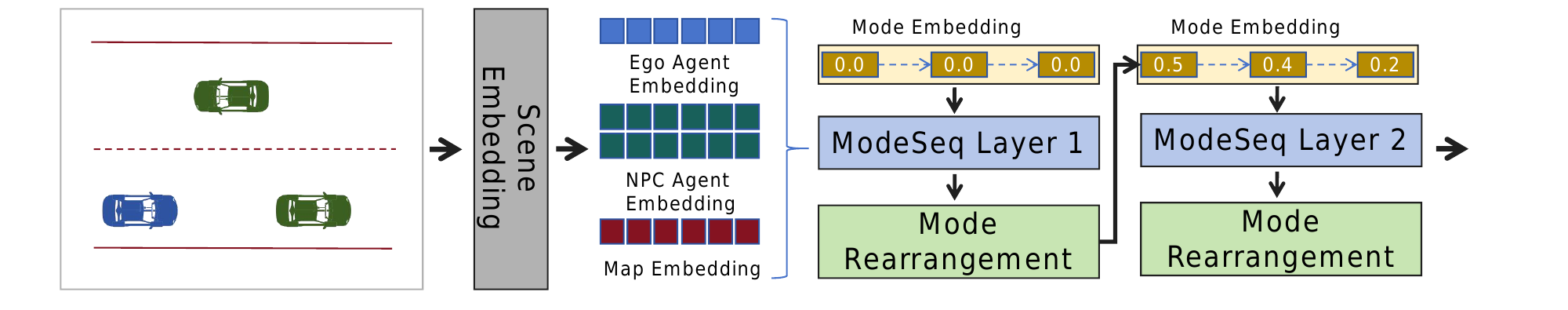}
    \caption{Overview of the Mode-as-Sequence framework. The scene embedding contains the ego agent's state, surrounding agent states, and map context, which are processed by multiple layers of sequential mode decoding. The output of each decoding layer is a sequence of mode embeddings, from which the corresponding trajectories and confidence scores can be predicted.}
    \label{fig:framework}
\end{figure*}

\paragraph{Generic Encoder-Decoder Architecture} 
Typical motion predictors adopt an encoder-decoder paradigm. An encoder first computes the scene embedding $\Psi$ from the input $S$. Based on $\Psi$, a decoder learns a set of latent embeddings $\{\mathbf{m}_{k}\}_{k=1}^{K}$ representing the $K$ future modes. Given the mode embedding $\mathbf{m}_k$, a prediction head outputs the trajectory and score via simple Multi-Layer Perceptrons (MLPs). This pipeline is summarized as:
\begin{equation}
\label{eq:generic_formulation}
\left\{
\begin{aligned}
    & \Psi = \text{Encoder}(S), \\
    & \{\mathbf{m}_{k}\}_{k=1}^{K} = \text{Decoder}(\Psi), \\
    & \hat{\mathbf{y}}_k, \hat{\phi}_k = \text{Head}(\mathbf{m}_{k}), \quad k \in \{1, \dots, K\}.
\end{aligned}
\right.
\end{equation}

\subsection{Motivation}
Multimodal trajectory forecasting is a fundamental task for autonomous driving systems, where the goal is to predict multiple possible future trajectories of the target agents given their past trajectories and the surrounding scene context. Traditional methods generate these trajectories as independent modes or hypotheses. However, these methods struggle to balance diversity (coverage of all plausible futures) and accuracy (matching the true future). Furthermore, they often fail to rank these modes effectively, which is critical for downstream decision-making in autonomous systems.

One of the main challenges in multimodal forecasting is that, during training, only one ground-truth future trajectory is available per scene, while multiple plausible future scenarios exist. This creates an inherent trade-off between mode diversity and mode accuracy. Existing methods, such as the winner-take-all (WTA) training strategy, supervise only one mode at a time and tend to focus on a single plausible future, neglecting other equally valid possibilities~\cite{alahi2016social, gupta2018social}.
On the other hand, confidence learning also poses a significant challenge. In many approaches, different modes are generated independently, leading to ambiguous confidence scores: it becomes difficult to assign confidence values that correctly reflect the likelihood of each mode since the model is not explicitly encouraged to capture the dependencies among modes. For instance, if the model generates two similar trajectories, both could have high confidence, even though one is a plausible continuation and the other is a redundant copy.

We address these challenges by transforming the multimodal trajectory forecasting problem into a sequence generation task. In Mode-as-Sequence, we treat the future modes as a \emph{sequence of dependent trajectories}, where each mode is generated conditioned on the previously generated modes. This formulation allows the model to capture the dependencies among different future scenarios, leading to better mode diversity and confidence ranking.
Under this framework, the prediction is no longer a set of independent modes, but a sequence of modes that are generated in consideration of their correlations. Such dependency structure naturally promotes the diversity by encouraging the model to generate modes that complement, rather than duplicate, the already generated trajectories. The sequential nature also paves a direct pathway toward effective mode ranking, where earlier modes are designed to represent the more likely futures. This paradigm helps mitigate the problem of mode collapse and improves the overall prediction quality.

While the sequential generation of modes in Mode-as-Sequence provides a natural way to capture dependencies and improve mode diversity, it suffers from efficiency limitations. Sequential generation requires generating the modes one by one, which can be computationally expensive, especially when a large number of modes is needed. For real-time applications, such as autonomous driving, this latency is a significant drawback.
To address this issue, we introduce Parallel ModeSeq, a parallelized version of Mode-as-Sequence. In Parallel ModeSeq, multiple modes are generated simultaneously, but the causal dependencies between modes are preserved through masked self-attention. This allows for efficient parallel generation of the modes while still maintaining their sequential structure and dependencies. By using causal self-attention, we ensure that each mode only depends on previously generated modes, allowing for parallelization without sacrificing accuracy or diversity.

The key advantage of Parallel ModeSeq over traditional parallel generation methods lies in its ability to preserve the causal structure between modes. In traditional parallel methods, such as those used in anchor-based or set-based generation, each mode is predicted independently, which can lead to redundant or inconsistent predictions. In contrast, Parallel ModeSeq maintains the natural dependencies between modes, ensuring that the generated trajectories form a coherent set of plausible futures.
Moreover, by generating the modes in parallel, Parallel ModeSeq significantly reduces the inference time compared with recurrent generation methods. This makes it particularly well-suited for real-time applications, where the ability to quickly predict multiple plausible futures is essential for decision-making and motion planning.

Overall, the Mode-as-Sequence framework, starting from a proof-of-concept recurrent version that demonstrates the potential of sequential mode modeling and transitioning to an efficient implementation that enjoys both the benefits of sequential generation and the efficiency of parallel decoding, forms a unified solution to addressing the challenges in multimodal motion prediction.

\subsection{Scene Encoding}
\label{sec:scene_encoding}
Since the focus of this work is on multimodal trajectory decoding, we mainly adopt QCNet~\cite{zhou2023query} as the scene encoder. QCNet has become one of the most effective scene encoding backbones in both academia and industry, largely due to its symmetric modeling of spatial and temporal structures with relative positional embeddings~\cite{shi2023mtr}. 

Specifically, QCNet first employs a hierarchical map encoding module based on map-map self-attention to produce map embeddings of shape $[M, D]$, where $M$ denotes the number of map elements and $D$ is the hidden dimension. In parallel, the agent encoder is built upon Transformer modules that explicitly factorize the spatial and temporal dimensions. These modules include temporal self-attention, agent-map cross-attention, and agent-agent self-attention. The three attention operations are organized as one group and interleaved twice, resulting in agent embeddings of shape $[A, T, D]$, where $A$ is the number of agents and $T$ is the number of historical time steps. Together, the agent embeddings and map embeddings constitute the final scene representation used by our decoder.

Although QCNet is adopted in this work for its strong empirical performance, our proposed framework is not restricted to a particular scene encoder. In principle, other scene encoding methods can also be incorporated into the proposed Mode-as-Sequence framework with moderate adaptation.
\begin{figure}[t]
  \centering
  \includegraphics[trim=0 300 400 0, clip,width=1\linewidth]{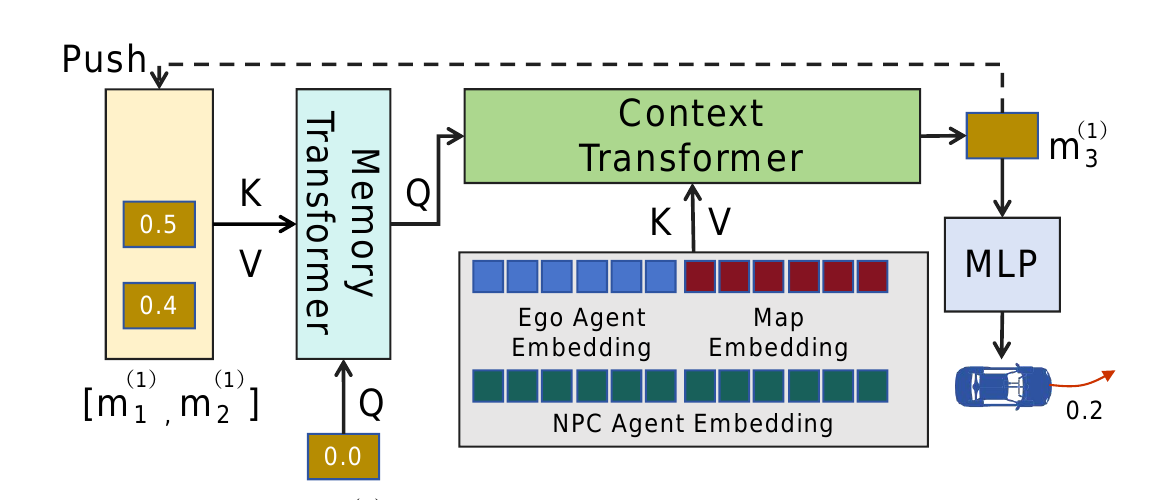}
    \caption{Details of ModeSeq layer. Each mode is iteratively predicted after looking at the previous modes in the Memory Transformer and fusing with the scene information in the Context Transformer.}
    \label{fig:modeseq_layer}
\end{figure}

\begin{figure}[t]
  \centering
  \includegraphics[trim=0 250 200 0, clip, width=1\linewidth]{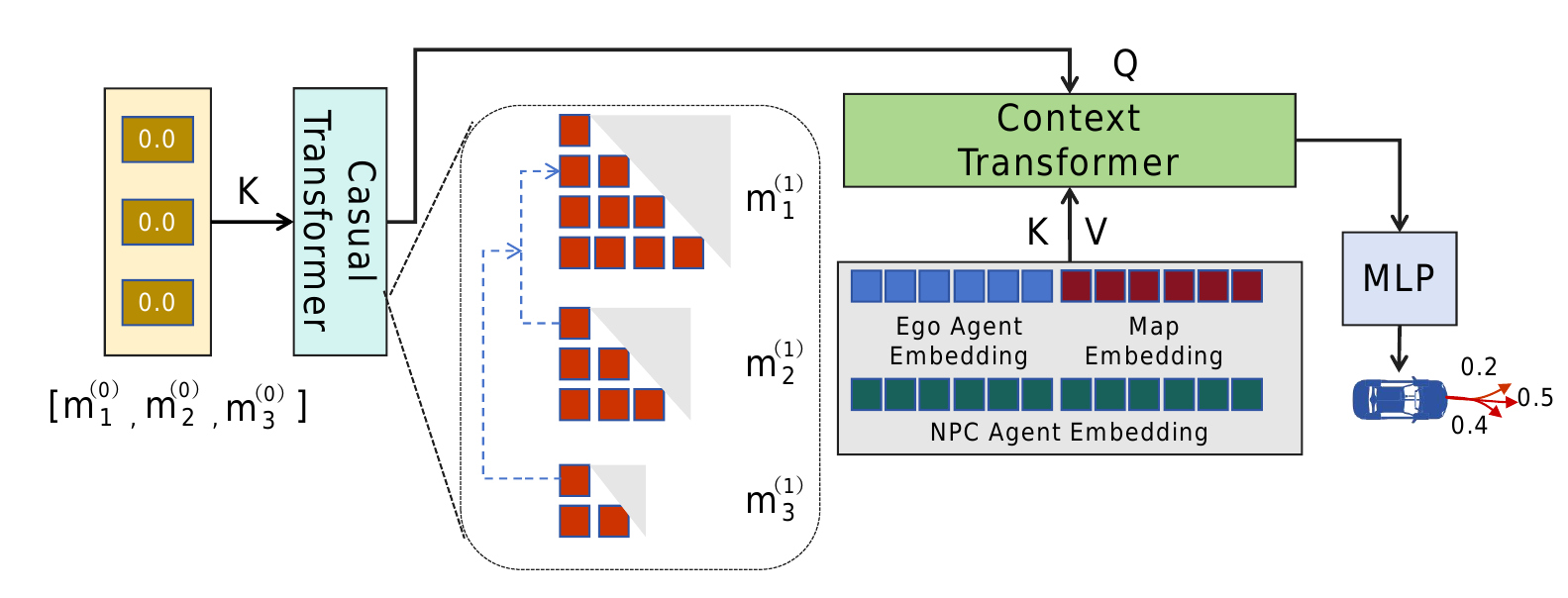}
    \caption{Details of Parallel ModeSeq layer. Multiple modes are predicted in parallel, with each mode conditioned on previous modes through causal self-attention. This allows for efficient parallel generation of modes while preserving dependencies between them.}
    \label{fig:parallel_modeseq_layer}
\end{figure}

\subsection{Single-Agent Mode Sequence Decoding}
\label{sec:single_agent_prediction}
We begin with the simplest setting of single-agent prediction, where the goal is to forecast a single agent's future trajectory conditioned on its past states and the surrounding scene context. Given the scene embedding $\Psi$ from the scene encoder, a trajectory decoder generates $K$ candidate future modes, each associated with an confidence score. Concretely, the pipeline can be summarized as
\begin{equation}
\quad
(m_1,\ldots,m_K)=\mathrm{Dec}_{\mathrm{seq}}(\Psi)
\end{equation}
where $m_k$ is the mode embedding, and $\mathrm{Dec}_{\mathrm{seq}}$ denotes a sequential mode decoder that produces an ordered mode sequence: modes are generated in the order of $k=1\rightarrow K$, and later modes are conditioned on the information already encoded in earlier modes. We instantiate two implementations of this same sequential decoding principle: ModeSeq realizes $\mathrm{Dec}_{\mathrm{seq}}$ via recurrent mode-by-mode decoding, while Parallel ModeSeq realizes $\mathrm{Dec}_{\mathrm{seq}}$ in one shot using causally masked mode-to-mode attention.

We instantiate two implementations of mode sequence decoding under the same factorization: ModeSeq (recurrent decoding) and Parallel ModeSeq (parallel decoding with causal dependency). Both aim to model mode-to-mode dependency, but differ in how the dependency is computed.

\paragraph{ModeSeq Decoder}
ModeSeq generates modes sequentially, predicting one mode after another. The key insight is that the $k$-th predicted mode should depend on previously predicted modes, which introduces an explicit inductive bias about mode dependency. At step $k$, the decoder takes the scene embedding $\Psi$ together with the set of previously generated mode embeddings $\{m_i\}_{i<k}$ and produces the next mode embedding. This sequential process helps the model avoid redundant hypotheses and makes confidence learning more structured, since later modes are conditioned on what has already been covered. Concretely, a ModeSeq layer can be written as:
\begin{equation}
\begin{aligned}
m_k
&= \text{CtxAttn}\!\Big(
    \text{MemAttn}(q_k,\{m_i\}_{i<k}),\ \Psi
\Big),\\
&\qquad k=1,\ldots,K .
\end{aligned}
\label{eq:seq_decoder}
\end{equation}
where $q_k$ is the learnable query for the $k$-th mode, $\text{MemAttn}(\cdot)$ performs mode-to-mode conditioning by reading previously generated modes, and $\text{CtxAttn}(\cdot)$ grounds the updated mode in the scene embedding $\Psi$.

\paragraph{Parallel ModeSeq Decoder}
Parallel ModeSeq addresses the efficiency limitation of recurrent decoding by generating all $K$ modes in one forward pass, while still preserving the causal mode dependency. Instead of looping over modes, it uses a causal attention mechanism along the mode dimension so that mode $k$ can only attend to modes $<k$. This retains the same dependency structure as ModeSeq, but is substantially more GPU-friendly and better suited when a larger number of modes is required. The core operation is causal mode-to-mode self-attention:
\begin{equation}
\mathrm{Attn}_{\mathrm{causal}}(\mathbf{Q},\mathbf{K},\mathbf{V})
=
\mathrm{Softmax}\!\left(\frac{\mathbf{Q}\mathbf{K}^{\top}}{\sqrt{D}}+\mathbf{M}_{\mathrm{causal}}\right)\mathbf{V}
\label{eq:causal_attn}
\end{equation}
where $\mathbf{Q},\mathbf{K},\mathbf{V}\in\mathbb{R}^{K\times D}$ denote the query/key/value matrices of the $K$ mode queries, and $\mathbf{M}_{\mathrm{causal}}$ is a lower-triangular mask (entries above the diagonal are set to $-\infty$), ensuring that mode $k$ only attends to modes $\le k$ (i.e., receives information from earlier modes). In our implementation, this is a mode-to-mode causal self-attention, hence $\mathbf{Q}=\mathbf{K}=\mathbf{V}$ by construction, but we write it in the standard QKV form for clarity.

\paragraph{Single-Layer Mode Sequence}
\label{sec:single_layer_modeseq}
This section describes the basic building block of ModeSeq: a single ModeSeq layer, as shown in Fig.~\ref{fig:modeseq_layer}. The layer decodes modes one-by-one and enforces mode dependency by explicitly reading previously decoded modes.

In a single ModeSeq layer, mode $k$ is produced by combining two sources of information: (i) the set of previous mode embeddings $\{m_i\}_{i<k}$ that summarize what futures have already been generated, and (ii) the scene embedding $\Psi$ that provides contextual constraints (map, interactions, etc.). Operationally, the layer can be viewed as:
\begin{itemize}
    \item First, update the current mode query by attending to previous modes (mode-to-mode conditioning).
    \item Then, condition the updated query on the scene embedding $\Psi$ (scene grounding).
    \item Finally, apply a lightweight prediction head to output $\hat{\mathbf{y}}_k$ and $\hat{c}_k$.
\end{itemize}
This design makes mode dependency explicit: later modes are generated with direct access to earlier modes, reducing duplicated predictions and stabilizing confidence ranking.

\paragraph{Single-Layer Parallel Mode Sequence}
\label{sec:single_layer_parallel_modeseq}
This section describes the basic building block of Parallel ModeSeq, illustrated in Fig.~\ref{fig:parallel_modeseq_layer}. The goal is to keep the same causal dependency as ModeSeq while removing the per-mode autoregressive loop.

Parallel ModeSeq starts from a set of learnable mode queries stacked as $\mathbf{Q}\in\mathbb{R}^{K\times D}$. A causal attention mask enforces that the $k$-th query only attends to $\{1,\ldots,k\}$, yielding parallel computation while preserving the Mode-as-Sequence factorization. A single Parallel ModeSeq layer can be summarized as:
\begin{itemize}
    \item Causal self-attention across modes enabled by a lower-triangular mask to propagate information from earlier modes to later ones.
    \item Scene conditioning via attention to $\Psi$ to ground each mode in the same scene context.
    \item Prediction head to output trajectories and confidences for all $K$ modes.
\end{itemize}

Compared with independent parallel heads (set prediction), causal masking makes the ordering meaningful and discourages later modes from ignoring what is already generated. Compared with recurrent ModeSeq, it eliminates the mode-by-mode loop and provides significantly faster inference, especially when $K$ is large.

\subsection{Multi-Agent Mode Sequence Decoding}
\label{sec:multi_agent_prediction}
While Sec.~\ref{sec:single_agent_prediction} focuses on forecasting a single target agent, practical autonomous driving requires predicting multiple interacting agents simultaneously. The key difficulty is that agent futures are not independent: plausible behaviors of one agent (e.g., yielding) may strongly constrain the futures of others (e.g., merging). A multi-agent predictor should therefore output joint futures that are interaction-consistent, diverse, and well-ranked.

Given the scene embedding $\Psi$ produced by the encoder, we consider the joint future motions of $A$ agents in the scene. The output is a set of $K$ joint modes. Each joint mode corresponds to one plausible future scene rollout, containing future trajectories for all the target agents:

\begin{equation}
\hat{\mathbf{Y}}_{k}=\{\hat{\mathbf{y}}_{k}^{a}\}_{a=1}^{A},\quad k\in\{1,\dots,K\},
\end{equation}
A scene-level confidence score $\hat{s}_k$ is also predicted to rank the joint futures.

A common strategy is to predict $K$ modes per agent and then compose them into joint futures. However, this leads to combinatorial explosion ($K^A$ joint candidates), interaction inconsistency (e.g., conflicting futures such as collisions), and unstable ranking (marginal confidences do not translate into a reliable scene-level ordering). We avoid these issues by predicting joint modes end-to-end and ranking them with a scene-level score.

Multi-agent prediction fits naturally into Mode-as-Sequence: instead of treating each agent independently, we decode a joint mode sequence where each mode represents a coherent multi-agent rollout. In practice, each joint mode $k$ maintains per-agent mode embeddings $\{m_k^{a}\}_{a=1}^{A}$ for the $A$ target agents, which are mapped to per-agent trajectories $\hat{\mathbf{y}}_{k}^{a}$ by lightweight heads. To explicitly model future interactions, we place a \emph{Future Interaction} module after the Context Transformer that performs self-attention among $\{m_k^{a}\}_{a=1}^{A}$ within the same joint scene, enabling information exchange across agents and producing interaction-consistent joint futures. To obtain a scene-level confidence for ranking joint modes, we aggregate the per-agent mode embeddings into a scene embedding via max pooling, $g_k=\mathrm{MaxPool}_{a}(m_k^{a})$, and then predict a scene score $\hat{s}_k=\mathrm{MLP}(g_k)$. This design keeps the output compact (only $K$ joint modes) while enforcing interaction consistency and providing a principled scene-level ranking signal.

Both instances of Mode-as-Sequence can be used for joint modes. In \textbf{Recurrent Joint ModeSeq}, joint modes are generated sequentially, each conditioned on previous joint modes, which provides strong coordination but incurs per-mode latency. In \textbf{Parallel Joint ModeSeq}, joint modes are generated in one shot with causal mode-to-mode attention, preserving joint-mode dependencies while enabling efficient inference. During training, each scene provides one ground-truth joint future (future trajectories for all agents). We optimize the model to produce at least one joint mode that matches the ground-truth joint future, assign a high scene-level score to that matched joint mode, and encourage other joint modes to cover alternative, interaction-consistent futures rather than duplicating the same rollout. The concrete objective is described in Sec.~\ref{sec:emta_training} with MA-EMTA and the ranking regularizer.

\subsection{Iterative Refinement with Multi-Layer Mode Sequences}
\label{sec:multi_layer_modeseq}
In this section, we extend the single-layer mode sequence decoder to a multi-layer architecture, which allows for iterative refinement of the generated modes. Multi-layer mode sequences help improve the accuracy and diversity of the predictions by progressively refining the mode predictions in each layer. We found that this approach can enhance the overall quality of the mode generation process, especially when the number of modes $K$ is large, and when more complex interactions are present in the scene.

\paragraph{Overview of Multi-Layer ModeSeq}
In the multi-layer ModeSeq architecture, we stack multiple mode decoding layers, each of which refines the mode embeddings generated in the previous layer. Each layer takes as input the mode embeddings from the previous layer and generates updated embeddings that represent the refined future trajectories. The number of layers $L$ is a hyperparameter, and the final prediction is made using the mode embeddings from the last layer.

At each layer $\ell$, we apply the following steps:
\begin{itemize}
    \item \textbf{Scene Context Update:} The mode embeddings from the previous layer, along with the scene embedding $\Psi$, are used to update the mode representations.
    \item \textbf{Mode Refinement:} The updated mode embeddings are refined through attention mechanism to incorporate the dependencies between modes and to better capture the interactions between the agents and the environment.
    \item \textbf{Mode Rearrangement:} We reorder the $K$ modes according to their predicted confidence scores, aligning the mode index with ranking before feeding them into the next layer.
\end{itemize}
This process continues until the final layer, where the final mode embeddings are used to predict the trajectories and confidence scores.

\paragraph{Mode Embeddings and Prediction Head}
Each mode embedding $m_k^{(\ell)}$ from layer $\ell$ is passed through a prediction head to output the corresponding future trajectory $\hat{\mathbf{y}}_k^{(\ell)}$ and its associated confidence score $\hat{c}_k^{(\ell)}$. The prediction head is typically a fully connected layer that maps the mode embedding to the trajectory space and applies a sigmoid function to generate the confidence score. The final predictions are made using the mode embeddings from the last layer $\ell = L$:
\begin{equation}
\hat{\mathbf{y}}_k, \hat{c}_k = \text{Head}(m_k^{(L)}), \quad k \in \{1, \dots, K\}.
\end{equation}

\paragraph{Advantages of Multi-Layer ModeSeq}
The multi-layer architecture provides several advantages over the single-layer decoder:
\begin{itemize}
 \item \textbf{Iterative Refinement:} Each layer allows the model to refine its predictions by incorporating feedback from previous modes and layers, improving prediction accuracy and diversity. This iterative process allows for more precise mode predictions, especially in complex scenarios.
 \item \textbf{Improved Mode Diversity:} By updating and refining the mode embeddings at each layer, the model is encouraged to explore a broader range of plausible futures. This is particularly useful in situations with high uncertainty or multiple competing behaviors.
 \item \textbf{Better Confidence Calibration:} As the model refines its predictions over multiple layers, the confidence scores for each mode become more calibrated, allowing for a more reliable ranking of the modes.
\end{itemize}

\paragraph{Training Multi-Layer ModeSeq}
Training the multi-layer ModeSeq model follows the same principles as the single-layer decoder, but with additional layer-wise supervision. Each layer $\ell$ is trained to produce better mode predictions and confidence scores. After each layer, we apply the Early-Match-Take-All (EMTA) loss and the ranking loss, as described in Sec.~\ref{sec:emta_training}, to ensure that the model produces accurate, diverse, and well-ranked modes. The overall training objective is an average of the layer-wise losses:
\begin{equation}
\mathcal{L}_{\text{multi-layer}} = \frac{1}{L}\sum_{\ell=1}^{L}  \mathcal{L}_{\text{layer}}^{(\ell)}.
\end{equation}

\paragraph{Extension to Multi-Agent Prediction}
The multi-layer ModeSeq decoder naturally extends to multi-agent joint-scene forecasting, where the output of each mode corresponds to a coherent multi-agent rollout. In this setting, the scene representation $\Psi$ encodes all agents and map context, and each joint mode maintains a set of per-agent mode embeddings $\{m_k^{a}\}_{a=1}^{A}$ for $A$ target agents. To explicitly model interactions, we place a \emph{Future Interaction} module after the Context Transformer at each decoding layer: it performs self-attention among the mode embeddings of all target agents within the same joint scene, allowing information exchange across agents and producing interaction-consistent joint futures. In addition, we introduce a \emph{scene scoring} module to obtain a scene-level confidence for ranking joint modes. Specifically, for each joint mode $k$, we aggregate the mode embeddings of all target agents into a scene embedding via max pooling, i.e., $g_k=\mathrm{MaxPool}_{a}(m_k^{a})$, and then project $g_k$ into a scene score $s_k$ using an MLP. Across layers, these interaction-aware refinements and scene-level scores enable iterative improvement of both joint trajectory quality and joint-mode ranking, which is critical for handling complex multi-agent interactions.

\subsection{Early-Match-Take-All Training}
\label{sec:emta_training}
In this section, we describe the training procedure for the Mode-as-Sequence framework using Early-Match-Take-All (EMTA) as the core training strategy. EMTA addresses the challenges of sparse supervision and mode collapse by enforcing the earliest match between the predicted modes and the ground truth, thus encouraging diverse and well-ranked predictions.

\paragraph{EMTA Label Assignment}
The key idea behind EMTA is to encourage the model to match the ground truth with the earliest predicted mode in the sequence. In traditional multimodal forecasting methods, a winner-take-all (WTA) approach is often used, where only the best-matched mode is supervised, leaving the other modes with insufficient guidance. This results in mode duplication and ambiguity in confidence ranking.

Instead, EMTA selects the \emph{earliest} mode that matches the ground truth and supervises it, while pushing the remaining modes to explore different plausible futures. This approach facilitates the diversity of the modes and provides a clear supervision signal for confidence learning.

Formally, we define the set of matched modes $\mathcal{M}$ as:
\begin{equation}
\mathcal{M} = \{ k \mid d(\hat{\mathbf{y}}_k, \mathbf{y}^{\star}) \le \delta \},
\end{equation}
where $d(\hat{\mathbf{y}}_k, \mathbf{y}^{\star})$ is a distance function (e.g., the displacement error of the trajectory endpoint or the average error over all waypoints), and $\delta$ is a threshold that determines whether a mode is considered a match to the ground truth $\mathbf{y}^{\star}$. The earliest matched mode is then selected as:
\begin{equation}
k^{\dagger} = \min \mathcal{M}.
\end{equation}
If no modes can match the ground truth, which frequently happens at the early stage of training, we fall back to the standard WTA selection:
\begin{equation}
k^{\dagger} = \arg\min_k d(\hat{\mathbf{y}}_k, \mathbf{y}^{\star}).
\end{equation}
The selected mode is then treated as the unique positive sample while all the remaining ones are deemed negative samples.

\paragraph{Regression and Classification}
Once the earliest matched mode $k^{\dagger}$ is selected, we apply a regression loss to optimize the model for getting closer to the ground-truth trajectory:
\begin{equation}
\mathcal{L}_{\text{reg}} = \mathcal{L}_{\text{NLL}}(\hat{\mathbf{y}}_{k^{\dagger}}, \mathbf{y}^{\star}),
\end{equation}
where $\mathcal{L}_{\text{NLL}}$ is a regression loss function based on the negative log-likelihood. Here, We follow the state-of-the-art approaches and use the negative log-likelihood of the parameterized Laplace distribution as the regression loss~\cite{zhou2023query, zhou2022hivt}. In addition to trajectory regression, we also optimize the confidence scores $\hat{c}_k$ for each mode. Since we desire the selected mode $k^{\dagger}$ to have a high confidence score and the remaining modes to have lower confidence, we define the ground truth of confidence scores as:
\begin{equation}
z_k =
\begin{cases}
1, & k = k^{\dagger}, \\
0, & \text{otherwise}.
\end{cases}
\end{equation}
The confidence scores are then optimized using the Binary Focal Loss $\mathcal{L}_{\text{focal}}$:
\begin{equation}
\mathcal{L}_{\text{cls}} = \frac{1}{K} \sum_{k=1}^{K} \mathcal{L}_{\text{focal}}(\hat{c}_k, z_k).
\end{equation}

\paragraph{Ranking Loss for Calibrated Confidences}
EMTA provides a simple supervision rule under sparse multimodal labels: among all predicted modes that are close enough to the ground truth, we supervise the earliest one in the mode sequence. This makes the role of each mode clearer---early modes are encouraged to be representative and high-confidence, while later modes are implicitly encouraged to explore alternatives instead of duplicating already-covered futures. In practice, this early-match principle yields more stable mode ordering and improves diversity without introducing complex auxiliary objectives.

However, when decoding is parallelized (Sec.~\ref{sec:single_layer_parallel_modeseq}) and especially when using multi-layer refinement with mode rearrangement, we observe that confidence inversions can still occur: a non-matched or lower-quality mode may be assigned a higher score than the matched mode. Such inversions directly hurt ranking-sensitive metrics and can destabilize the intended ``early-is-better'' mode ordering. To strengthen confidence consistency, we add a lightweight margin-based ranking loss on top of EMTA.

Concretely, let $k^{\dagger}$ denote the EMTA-selected mode (earliest match, with WTA fallback when no match exists). We enforce that the matched mode should have a higher confidence than the remaining modes by a margin. This is implemented with a hinge ranking term that penalizes confidence inversions:
\begin{equation}
\mathcal{L}_{\text{rank}} = \frac{1}{|\mathcal{N}|} \sum_{k \in \mathcal{N}} \max\left(0, \gamma - (\hat{c}_{k^{\dagger}} - \hat{c}_k)\right),
\end{equation}
where $\mathcal{N}$ is the set of negative modes that do not match the ground truth, and $\gamma$ is a margin hyperparameter that enforces a minimum confidence difference between the matched mode and the non-matched ones, which is set to $0.1$ by default. This objective is intentionally simple: it does not change which mode is regressed to the ground truth, but it makes the confidence ordering more robust---particularly important for Parallel ModeSeq where all modes are produced in one shot and the model must learn a consistent global ordering.

\paragraph{Overall EMTA Training Objective}
The overall loss function for EMTA training is a weighted sum of the regression loss, confidence loss, and ranking loss:
\begin{equation}
\mathcal{L}_{\text{EMTA}} = \mathcal{L}_{\text{reg}} + \lambda_{\text{cls}} \mathcal{L}_{\text{cls}} + \lambda_{\text{rank}} \mathcal{L}_{\text{rank}},
\end{equation}
where $\lambda_{\text{cls}}$ and $\lambda_{\text{rank}}$ are hyperparameters that control the relative importance of the confidence and ranking losses.


\paragraph{Multi-Agent EMTA}
\label{sec:ma_emta}
For multi-agent forecasting, the supervision is naturally scene-level: each training scene provides one ground-truth joint future (future trajectories for all agents). Instead of supervising each agent independently and then composing marginal predictions, we train joint modes directly, as introduced in Sec.~\ref{sec:multi_agent_prediction}. Each joint mode represents a coherent multi-agent rollout and is associated with a scene-level score used for ranking joint futures.
Multi-Agent EMTA (MA-EMTA) extends the early-match principle from single-agent modes to joint modes. The key idea remains the same: among joint modes that match the ground-truth joint future, we supervise the earliest one in the mode sequence. This produces a single, stable target per scene for both regression and scene-level ranking, while encouraging later joint modes to cover alternative interaction-consistent futures.

To determine whether a joint mode matches the ground truth, we compute a scene-level distance between the predicted joint rollout and the ground-truth joint rollout, e.g., by aggregating per-agent distances. Let $D(\hat{\mathbf{Y}}_k,\mathbf{Y}^{\star})$ denote this distance and let $\Delta$ be the matching threshold. MA-EMTA selects the earliest matched joint mode; if no joint mode matches, it falls back to the closest joint mode under $D(\cdot)$.
We then regress all agents under the selected joint mode and supervise the scene-level score so that the selected joint mode is ranked above the others.

Similar to the single-agent case, we also apply the same ranking regularizer at the scene level to prevent confidence inversions between joint modes. This is particularly important when joint modes are decoded in parallel, where score consistency is crucial for selecting a reliable scene rollout.

Overall, MA-EMTA provides an efficient way to learn diverse and interaction-consistent joint futures with well-ranked scene-level scores, without combinatorial enumeration of marginal modes.

\section{Experiments}
\label{sec:experiments}

\subsection{Experimental Setup}
\label{sec:experimental_setup}
\begin{table*}[t]
\centering
\caption{Quantitative results on the 2024 Waymo Open Dataset Motion Prediction Benchmark. Results are reported for $K=6$ modes.}
\label{tab:womd_results}
\resizebox{\linewidth}{!}{
\begin{tabular}{@{}lcccccccc@{}}
\toprule
Dataset & Method & Ensemble & Lidar & Soft mAP $\uparrow$ & mAP $\uparrow$ & MR $\downarrow$ & minADE $\downarrow$ & minFDE $\downarrow$ \\ \midrule
\multirow{4}{*}{Val} & MTR v3~\cite{maptrv3} & $\times$ & \checkmark & - & 0.4593 & 0.1175 & 0.5791 & 1.1809 \\
 & MTR++~\cite{shi2023mtr} & $\times$ & $\times$ & - & 0.4382 & 0.1337 & 0.6031 & 1.2135 \\
 & QCNet~\cite{zhou2023query} & $\times$ & $\times$ & 0.4508 & 0.4452 & 0.1254 & 0.5122 & 1.0225 \\
 &ModeSeq (Ours) & $\times$ & $\times$ &0.4562 &0.4507 &0.1206 &0.5237 &1.0681 \\ \midrule
\multirow{4}{*}{Test} & MTR v3~\cite{maptrv3} & \checkmark & \checkmark & 0.4967 & 0.4859 & 0.1098 & 0.5554 & 1.1062 \\
 & RMP Ensemble~\cite{rmp} & \checkmark & $\times$ & 0.4726 & 0.4553 & 0.1113 & 0.5596 & 1.1272 \\
 & ModeSeq (Ours) & \checkmark & $\times$ & 0.4737 & 0.4665 & 0.1204 & 0.5680 & 1.1766 \\
 & ModeSeq (Ours) & $\times$ & $\times$ & 0.4487 & 0.4450 & 0.1244 & 0.5304 & 1.0836 \\ \bottomrule
\end{tabular}
}
\end{table*}
\paragraph{Datasets} 
We evaluate the proposed Mode-as-Sequence framework on the Waymo Open Motion Dataset (WOMD)~\cite{ettinger2021large}, which is currently the most challenging large-scale benchmark for motion forecasting. The dataset contains over 500,000 segments collected from diverse urban driving scenarios. For the marginal prediction task, we utilize the standard training, validation, and testing splits. For the interactive prediction task, we follow the protocol of the 2025 Waymo Interactive Prediction Challenge, which requires predicting the future joint distributions of two interacting agents of interest. We additionally evaluate on Argoverse 2~\cite{wilson2021argoverse}, a large-scale autonomous driving benchmark with diverse urban scenarios and high-definition maps; we follow the official split and reporting protocol for single-agent forecasting (e.g., $K{=}6$ modes) to assess the generalization of Mode-as-Sequence beyond WOMD.

\paragraph{Metrics} 
Following the standard evaluation protocol of WOMD, we report the average displacement error (ADE), final displacement error (FDE), and miss rate (MR). To evaluate the quality of the predicted multimodal distribution and confidence calibration, we primarily focus on the mean average precision (mAP), which is the official ranking metric for the challenge. In the multi-agent joint prediction track, we additionally report the joint-mAP, joint-ADE, and joint-FDE to measure the coordination between agents.

\paragraph{Implementation Details} 
We adopt QCNet~\cite{zhou2023query} as the default scene encoder, producing embeddings with a hidden dimension of $D=128$. The decoder consists of $L=6$ stacked ModeSeq layers. In each layer, the number of modes is set to $K=6$ for marginal prediction and $K=6$ for joint prediction (representing scene-level modes). The Mode-to-Mode Transformer and Context Transformer are both implemented with 8 attention heads. For the ranking loss, the margin $\gamma$ is empirically set to $0.1$.

\begin{table}[t]
\centering
\caption{Quantitative results on the Argoverse 2 Single-Agent Motion Forecasting Benchmark. Results are reported for $K=6$ modes.}
\label{tab:av2_results}
\resizebox{\linewidth}{!}{
\begin{tabular}{@{}lccccc@{}}
\toprule
Method & Ensemble & b-minFDE $\downarrow$ & MR $\downarrow$ & minADE $\downarrow$ & minFDE $\downarrow$ \\ \midrule
MTR~\cite{maptrv3} & \checkmark & 1.98 & 0.15 & 0.73 & 1.44 \\
MTR++~\cite{shi2023mtr} & \checkmark & 1.88 & 0.14 & 0.71 & 1.37 \\
QCNet~\cite{zhou2023query} & $\times$ & 1.91 & 0.16 & 0.65 & 1.29 \\
\textbf{ModeSeq (Ours)} & $\times$ & \textbf{1.87} & \textbf{0.14} & \textbf{0.63} & \textbf{1.26} \\ \bottomrule
\end{tabular}
}
\end{table}

\paragraph{Training Protocol} 
The framework is implemented in PyTorch and trained on 4 NVIDIA L20 GPUs. We use the AdamW optimizer with a weight decay of $0.1$, a dropout rate of $0.1$, and an initial learning rate of $5 \times 10^{-4}$ decayed following a cosine annealing schedule. The total training duration is $30$ epochs with a total batch size of $32$.

\begin{table*}[t]
\centering
\small
\setlength{\tabcolsep}{6pt}
\caption{Quantitative results on the 2025 Waymo Open Dataset Interaction Prediction Benchmark. Results are reported for $K=6$ modes at the scene level, involving the joint future of $2$ to $8$ agents.}
\label{tab:womd_leaderboard_avg_sorted_fde_desc}
\resizebox{\linewidth}{!}{
\begin{tabular}{l|ccccc}
\toprule
Method & Min ADE $\downarrow$ & Min FDE $\downarrow$ & Miss Rate $\downarrow$ & mAP v2 $\uparrow$ & Soft-mAP v2 $\uparrow$ \\
\midrule
Waymo LSTM baseline~\cite{waymo}      & 1.9056 & 5.0278 & 0.7750 & 0.0648 & 0.0693 \\
infgen-base-xl           & 1.2059 & 2.6427 & 0.5649 & 0.0714 & 0.1236 \\
Causal-Planner++~\cite{yuan2025causal}         & 0.9845 & 2.2958 & 0.4338 & 0.2499 & 0.2557 \\
BeTop-ens~\cite{liu2024reasoning}                & 0.9779 & 2.2805 & 0.4376 & 0.2511 & 0.2573 \\
IMPACT~\cite{sun2025impact}                   & 0.9738 & 2.2734 & 0.4316 & 0.2659 & 0.2718 \\
AutoDiffuser-Draft       & 0.9422 & 2.1759 & 0.4885 & 0.2211 & 0.2249 \\
TrajFlow~\cite{yan2025trajflow}                 & 0.9381 & 2.1521 & 0.4180 & 0.2533 & 0.2593 \\
RMP YOLO~\cite{sun2025rmp}                 & 0.9274 & 2.1131 & 0.4167 & 0.2313 & 0.2486 \\
qcnet R braids           & 0.9156 & 2.0946 & 0.4801 & 0.1490 & 0.1922 \\
RetroMotion (SMoE hybrid)~\cite{wagner2025retromotion}& 0.9256 & 2.0890 & 0.4347 & 0.2519 & 0.2562 \\
\textbf{Parallel ModeSeq(Ours)}         & \textbf{0.7707} & \textbf{1.6897} & \textbf{0.3782} & \textbf{0.2949} & \textbf{0.2978} \\
\bottomrule
\end{tabular}
}
\end{table*}
\begin{table*}[h]
\centering
\caption{Effects of sequential mode modeling and Early-Match-Take-All training on the validation set of the WOMD. All models adopt the QCNet encoder.}
\label{tab:ablation_components}
\resizebox{\linewidth}{!}{
\begin{tabular}{@{}l|lcccccc@{}}
\toprule
Decoder & Training Strategy & Ignored Samples & Soft $mAP_6 \uparrow$ & $mAP_6 \uparrow$ & $MR_6 \downarrow$ & minADE$_6 \downarrow$ & minFDE$_6 \downarrow$ \\ \midrule
DETR w/Refinement & WTA & None & 0.4096 & 0.4050 & 0.1536 & 0.5660 & 1.1716 \\
DETR w/Refinement & WTA & Other Matches & 0.4150 & 0.4103 & 0.1502 & 0.5619 & 1.1621 \\ \midrule
Mode-as-Sequence (Ours) & WTA & None & 0.4138 & 0.4093 & 0.1502 & 0.5563 & 1.1498 \\
Mode-as-Sequence (Ours) & WTA & Other Matches & 0.4207 & 0.4161 & 0.1503 & 0.5556 & 1.1501 \\ \midrule
Mode-as-Sequence (Ours) & EMTA & None & \textbf{0.4231} & \textbf{0.4196} & \textbf{0.1457} & 0.5700 & 1.1851 \\
Mode-as-Sequence (Ours) & EMTA & Other Matches & 0.4098 & 0.4060 & 0.1496 & 0.5817 & 1.2207 \\ \bottomrule
\end{tabular}
}
\end{table*}

\subsection{Comparison with State of the Art}
\label{sec:exp_sota}
We compare ModeSeq with representative state-of-the-art motion forecasting systems on the Waymo Open Dataset (WOMD) motion prediction benchmark and Argoverse 2 single-agent forecasting. Following the benchmarks, we report Soft mAP / mAP as the primary ranking metrics (higher is better), together with miss rate (MR), minADE, and minFDE (lower is better). We additionally annotate whether each method uses model ensembling (Ensemble) or LiDAR point clouds (LiDAR).

\paragraph{WOMD Motion Prediction Benchmark}
Table~\ref{tab:womd_results} compares ModeSeq with strong public baselines on WOMD with explicit Ensemble/LiDAR annotations. On the LiDAR-free validation split, ModeSeq achieves the best ranking performance among LiDAR-free methods, improving Soft mAP/mAP over QCNet (0.4508/0.4452 $\rightarrow$ 0.4562/0.4507) and reducing MR (0.1254 $\rightarrow$ 0.1206), suggesting that sequential mode decoding yields a more representative and better-scored hypothesis set under sparse supervision. On the test split, ModeSeq remains strong in the single-model setting (0.4487/0.4450) and further benefits from ensembling without LiDAR (0.4737/0.4665), outperforming RMP Ensemble in ranking metrics (0.4737/0.4665 vs.\ 0.4726/0.4553). Although MTR v3 uses both ensembling and LiDAR, ModeSeq narrows the gap substantially while staying LiDAR-free, highlighting the effectiveness of Mode-as-Sequence decoding for high-quality sparse multimodal prediction.

\paragraph{Argoverse 2 Single-Agent Forecasting Benchmark}
Table~\ref{tab:av2_results} reports Argoverse 2 single-agent results with $K=6$ modes. Without ensembling, ModeSeq achieves the best overall accuracy, improving over QCNet in b-minFDE (1.91 $\rightarrow$ 1.87), MR (0.16 $\rightarrow$ 0.14), and minADE/minFDE (0.65/1.29 $\rightarrow$ 0.63/1.26), indicating better coverage and best-of-$K$ precision under the same sparse supervision. Despite MTR and MTR++ using ensembles, ModeSeq remains competitive and attains the lowest b-minFDE (1.87 vs.\ 1.88/1.98) while matching the best miss rate (0.14), suggesting that mode-as-sequence decoding generalizes well beyond WOMD without relying on ensembling.

\paragraph{WOMD Interaction Prediction Benchmark}
Table~\ref{tab:womd_leaderboard_avg_sorted_fde_desc} reports a leaderboard snapshot under a fixed evaluation setting (object type = All, measurement time = Avg), where each entry summarizes performance aggregated across horizons and classes. Although the table is sorted by Min FDE in descending order to illustrate the metric range (lower is better), Parallel ModeSeq stands out by achieving the best results simultaneously on ranking-based metrics and displacement/miss metrics: it attains the highest mAP v2 / Soft-mAP v2 (0.2949 / 0.2978) while also obtaining the lowest minADE / minFDE / Miss Rate (0.7707 / 1.6897 / 0.3782). Compared with strong recent baselines in the same snapshot, Parallel ModeSeq improves mAP v2 by +0.0231 over IMPACT (0.2659 $\rightarrow$ 0.2949) and by +0.0416 over TrajFlow (0.2533 $\rightarrow$ 0.2949), while reducing minFDE substantially (2.2734 $\rightarrow$ 1.6897, a 25.7\% decrease vs.\ IMPACT; 2.1521 $\rightarrow$ 1.6897, a 21.5\% decrease vs.\ TrajFlow) and lowering miss rate (0.4316 $\rightarrow$ 0.3782, down 0.0534 vs.\ IMPACT; 0.4180 $\rightarrow$ 0.3782, down 0.0398 vs.\ TrajFlow). The fact that Parallel ModeSeq improves mAP-style ranking and minFDE/MR at the same time suggests that our approach enhances both the quality of the predicted trajectories and the reliability of confidence scoring, rather than trading one for the other.
\begin{table*}[t]
\centering
\caption{Effects of mode rearrangement on the validation set of the WOMD. These experiments evaluate the synergy between reordering and different label assignment variants.}
\label{tab:ablation_rearrangement}
\resizebox{\linewidth}{!}{
\begin{tabular}{@{}cc|ccccc@{}}
\toprule
Mode Rearrangement & Ignored Samples & Soft $mAP_6 \uparrow$ & $mAP_6 \uparrow$ & $MR_6 \downarrow$ & minADE$_6 \downarrow$ & minFDE$_6 \downarrow$ \\ \midrule
$\times$ & None & 0.4112 & 0.4077 & 0.1548 & 0.5884 & 1.2389 \\
$\times$ & Early Mismatches & 0.4141 & 0.4109 & 0.1489 & 0.5749 & 1.2066 \\ \midrule
\checkmark & None & \textbf{0.4231} & \textbf{0.4196} & \textbf{0.1457} & \textbf{0.5700} & \textbf{1.1851} \\
\checkmark & Early Mismatches & 0.4161 & 0.4129 & 0.1461 & 0.5751 & 1.2041 \\ \bottomrule
\end{tabular}
}
\end{table*}

\subsection{Ablation Study}
\label{sec:ablation_study}
In this section, we conduct a series of ablation experiments on the WOMD validation set to verify the effectiveness of each core component in the Parallel Mode-as-Sequence framework. All models are trained with a QCNet encoder for 30 epochs unless otherwise specified.
\paragraph{Effects of Sequential Mode Modeling}
In Table~\ref{tab:ablation_components}, we examine the effectiveness of sequential mode modeling by comparing Mode-as-Sequence with a sparse DETR-like decoder enhanced with iterative refinement, both employing the same QCNet encoder~\cite{zhou2023query} for fair comparisons. The results demonstrate that Mode-as-Sequence outperforms the baseline on all metrics when using the same training strategy. Interestingly, ignoring the confidence loss of the suboptimal modes that match the ground truth can improve the performance of both methods under the WTA training. The reason behind this is that treating the other matched modes as negative samples will confuse the optimization process, given that the best and the other matches usually have similar mode representations while they are assigned as opposite samples.

\begin{figure}[t]
  \centering
  \includegraphics[width=1.0\linewidth]{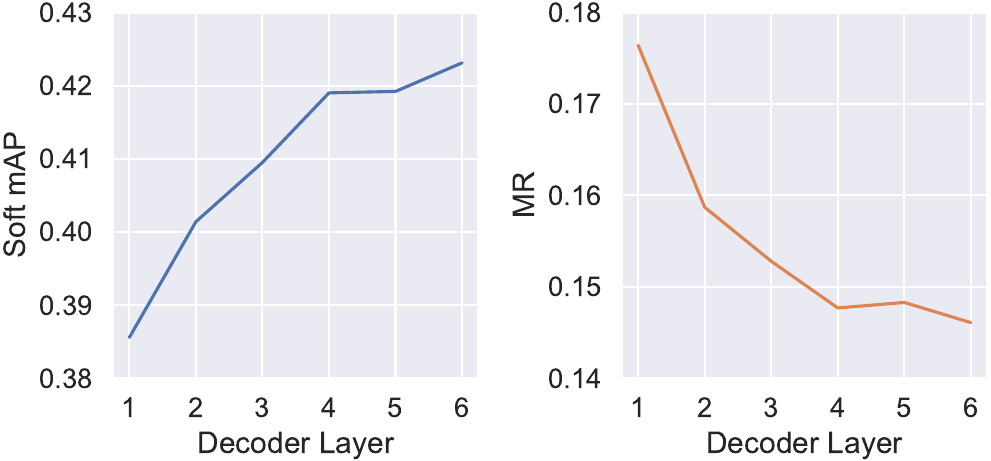}
  \vspace{-0.7cm}
  \caption{The performance after each decoding layer on the validation set of the WOMD.}
  \label{fig:refinement}
  \vspace{-0.7cm}
\end{figure}
\paragraph{Effects of EMTA Training}
We also investigate the role of EMTA training in Table~\ref{tab:ablation_components}. After replacing the WTA loss with our EMTA scheme, the results on Soft mAP${}_6$, mAP${}_6$, and MR${}_6$ are considerably improved, which demonstrates the benefits of EMTA training in terms of mode coverage and confidence scoring. On the other hand, the performance on minADE${}_6$ and minFDE${}_6$ slightly deteriorates since the EMTA loss has relaxed the requirement for trajectory accuracy, but the degree of deterioration is less than $0.02$ meters in minADE${}_6$ and $0.04$ meters in minFDE${}_6$, leading to more balanced performance taken overall. Contrary to the conclusion drawn from the WTA baselines, treating other matches as ignored samples is detrimental under the EMTA strategy. This is because the joint effects of sequential mode decoding and EMTA training have broken the symmetry of mode modeling and label assignment, allowing us to assign the other matches as negative samples to drive them away from the ground truth for covering other likely modes.

\begin{figure*}[t]
\centering
\subfloat[\scriptsize \#Mode@Training=3, \#Mode@Inference=3\label{fig:womd_mode_3_3}]{
  \includegraphics[width=0.31\linewidth]{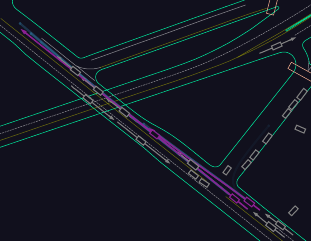}
}\hfill
\subfloat[\scriptsize \#Mode@Training=6, \#Mode@Inference=6\label{fig:womd_mode_6_6}]{
  \includegraphics[width=0.31\linewidth]{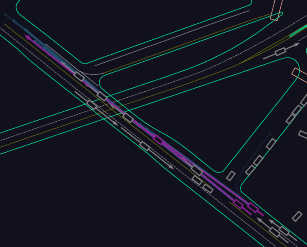}
}\hfill
\subfloat[\scriptsize \#Mode@Training=6, \#Mode@Inference=24\label{fig:womd_mode_6_24}]{
  \includegraphics[width=0.31\linewidth]{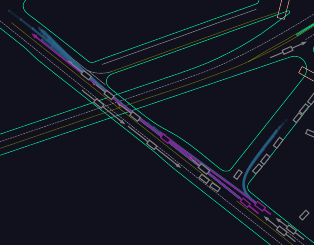}
}
\caption{Visualization on the WOMD. The agents in purple are predicted with blue trajectories, with the opacity indicating confidence.}
\label{fig:womd_mode_extrapolation}
\end{figure*}

\paragraph{Effects of Iterative Refinement}
To understand the role of iterative refinement in our framework, we evaluate the outputs after each decoding layer. As shown in Fig.~\ref{fig:refinement}, deeper decoding generally improves both ranking quality and coverage: Soft mAP${}_6$ increases steadily across layers, while MR${}_6$ decreases accordingly, indicating that refinement not only sharpens trajectory hypotheses but also makes the confidence ordering more reliable for selecting representative futures. We observe the largest gains in the early layers, followed by diminishing returns as the decoder converges, which suggests that the refinement process quickly corrects coarse hypotheses and then focuses on fine-grained adjustments. Intuitively, each additional layer can revisit the mode embeddings with updated context, resolve ambiguities left by earlier decoding steps, and reduce redundancy among modes, leading to better mode separation. A key reason why iterative refinement is particularly effective in Mode-as-Sequence is the mode rearrangement operation inserted between layers: by reordering modes according to their predicted confidence, the next layer prioritizes refining the more plausible trajectories first and pushes low-confidence or noisy modes toward the tail of the sequence. This reduces the negative impact of early bad modes on subsequent decoding and aligns well with EMTA’s preference for early high-quality matches, together yielding consistent improvements in both Soft mAP and MR.

\paragraph{Effects of Mode Rearrangement}
Mode rearrangement coordinates with EMTA training to facilitate decoding matched trajectories with high confidence as early as possible. Comparing the corresponding settings in Table~\ref{tab:ablation_rearrangement}, we can see that reordering the mode embeddings before further refinement can remarkably promote the forecasting capability. To gain deeper insights into the results, we develop a variant of label assignment, where the modes decoded earlier than the first match are deemed ignored samples. Again, incorporating mode rearrangement is beneficial under this setting. Interestingly, we found this strategy of label assignment to outperform the default one in the absence of mode rearrangement. This phenomenon can be explained by the fact that bad modes may appear in the first few decoding steps of the shallow layers, which can negatively impact the learning of the subsequent modes. To implicitly guide the model to output more confident modes first, we can provide monotonically decreasing confidence labels by ignoring the confidence loss of the early mismatches. Playing a similar but stronger role, mode rearrangement explicitly prioritizes the refinement of the more probable trajectories in the next layer by manually placing the less confident modes at the end of the sequence.

\begin{table}[t]
\centering
\caption{Capability of generating representative modes with precise confidence scores and corresponding inference latency. Models are evaluated on the WOMD validation split.}
\label{tab:capability_latency}
\resizebox{\linewidth}{!}{
\begin{tabular}{@{}cl|cccc@{}}
\toprule
\#Mode & Model & Soft $mAP_6 \uparrow$ & $mAP_6 \uparrow$ & $MR_6 \downarrow$ & Latency (ms) \\ \midrule
\multirow{2}{*}{3} & QCNet [51] & 0.4214 & 0.4163 & 0.2007 & $63\pm11$ \\
 & Mode-as-Sequence (Ours) & 0.4509 & 0.4479 & 0.1967 & $86\pm9$ \\ \midrule
\multirow{2}{*}{6} & QCNet [51] & 0.4508 & 0.4452 & 0.1254 & $69\pm16$ \\
 & Mode-as-Sequence (Ours) & \textbf{0.4562} & \textbf{0.4507} & \textbf{0.1206} & $143\pm10$ \\ \bottomrule
\end{tabular}
}
\end{table}
\begin{figure}[t]
  \centering
  \includegraphics[width=1.0\linewidth]{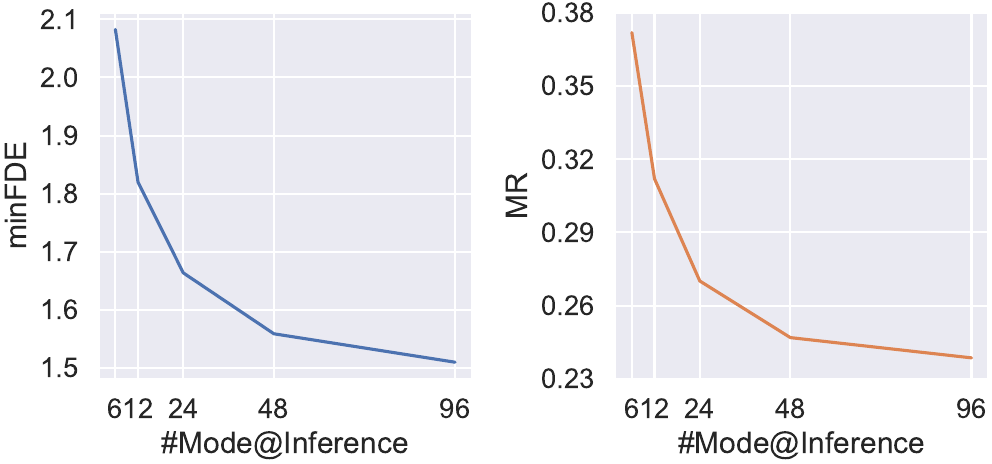}
  \vspace{-0.2cm}
  \caption{The results of generating more than $6$ modes on the validation set of the WOMD.}
  \label{fig:extrapolation}
  \vspace{-0.6cm}
\end{figure}

\paragraph{Capability of Representative Mode Learning.}
We demonstrate Mode-as-Sequence's ability to produce representative modes in Table~\ref{tab:capability_latency}. While training models to decode merely $3$ modes necessarily leads to worse performance in general, the $3$-mode variant of Mode-as-Sequence achieves a comparable level of performance on Soft mAP${}_6$ and mAP${}_6$ to the $6$-mode model, indicating that the early decoded modes capture the most informative and frequent futures rather than collapsing to redundant hypotheses. This behavior is consistent with the mode-as-sequence inductive bias: later modes are generated conditioned on earlier ones, so the model is encouraged to allocate capacity to diverse behaviors instead of duplicating the same trajectory. Practically, this result implies that Mode-as-Sequence can be trained and deployed with a smaller decoding budget while still maintaining strong ranking quality, which is attractive for latency-sensitive systems where downstream planning often uses only a few top hypotheses. By comparison, QCNet~\cite{zhou2023query} fails to achieve comparable results if only using $3$ mode queries during training, suggesting that set-style decoders are more sensitive to the number of queries and tend to lose coverage/ranking quality when $K$ is reduced.

\paragraph{Capability of Mode Extrapolation.}
We ask the model trained by generating $6$ modes to execute more decoding steps at test time. As depicted in Fig.~\ref{fig:extrapolation}, Mode-as-Sequence achieves consistently lower prediction error as the number of decoded modes increases, exhibiting a clear mode extrapolation capability enabled by sequential modeling. Intuitively, additional decoding steps allow the model to place new hypotheses conditioned on already generated ones, which helps fill remaining plausible futures rather than repeating existing trajectories. The curves also show diminishing returns: the largest gain is obtained when increasing from a small number of modes to a moderate number, after which improvements become more gradual, reflecting that the most likely behaviors are already covered by early modes. This extrapolation property is particularly useful in deployment, since the required number of hypotheses varies with scene ambiguity; the same trained model can decode more modes on-demand for complex interactions while keeping a smaller budget for simple scenes, without retraining or changing the encoder.  

\paragraph{Inference Cost}
As shown in Table~\ref{tab:capability_latency}, the inference latency of ModeSeq is about twice as high as that of QCNet~\cite{zhou2023query} if predicting $6$ modes. However, in many industrial autonomous driving solutions, the number of modes used by downstream planning is refrained from exceeding $3$. Table~\ref{tab:capability_latency} shows that the latency gap between the $3$-mode models is much smaller, while the mAP${}_6$ of $3$-mode ModeSeq is even better than that of $6$-mode QCNet. Given our approach's capability of producing representative trajectories with fewer modes, we believe sequential mode modeling has the potential to be deployed on board.

Distinct from the original Recurrent ModeSeq, the Parallel ModeSeq framework enables simultaneous mode generation through causal self-attention masks. This modification allows the model to maintain the high trajectory diversity and calibrated confidence of sequential modeling while achieving a competitive inference latency. In our experiments, the parallelized decoder achieves an average inference time of approximately $48$ms per scene, making it suitable for real-time deployment on autonomous driving platforms.

\begin{table*}[t]
\centering
\small
\setlength{\tabcolsep}{6pt}
\caption{WOMD leaderboard breakdown (ModeSeq, Motion Prediction): Avg-only results over horizons.}
\label{tab:modeseq_avg_only}
\resizebox{\linewidth}{!}{
\begin{tabular}{l|cccccc}
\toprule
Type & Soft mAP v2 $\uparrow$ & mAP v2 $\uparrow$ & Min ADE $\downarrow$ & Min FDE $\downarrow$ & Miss Rate $\downarrow$ & Overlap Rate $\downarrow$ \\
\midrule
Vehicle    & 0.5181 & 0.5095 & 0.6780 & 1.4046 & 0.1129 & 0.0465 \\
Pedestrian & 0.4781 & 0.4709 & 0.3407 & 0.7117 & 0.0776 & 0.2617 \\
Cyclist    & 0.4248 & 0.4190 & 0.6852 & 1.4136 & 0.1706 & 0.0743 \\
All        & 0.4737 & 0.4665 & 0.5680 & 1.1766 & 0.1204 & 0.1275 \\
\bottomrule
\end{tabular}}
\end{table*}
\begin{table*}[t]
\centering
\small
\setlength{\tabcolsep}{6pt}
\caption{WOMD leaderboard breakdown (Parallel ModeSeq, Interaction Prediction): Avg-only results over horizons.}
\label{tab:parallel_modeseq_avg_only}
\resizebox{\linewidth}{!}{
\begin{tabular}{l|ccccccc}
\toprule
Type & Soft mAP $\uparrow$ & mAP $\uparrow$ & Min ADE $\downarrow$ & Min FDE $\downarrow$ & Miss Rate $\downarrow$ & mAP v2 $\uparrow$ & Overlap Rate $\downarrow$ \\
\midrule
Vehicle    & 0.4107 & 0.4063 & 0.7864 & 1.6866 & 0.2843 & 0.4063 & 0.1328 \\
Pedestrian & 0.3059 & 0.3029 & 0.6093 & 1.3241 & 0.3598 & 0.3029 & 0.2909 \\
Cyclist    & 0.1768 & 0.1754 & 0.9164 & 2.0585 & 0.4904 & 0.1754 & 0.1451 \\
All        & 0.2978 & 0.2949 & 0.7707 & 1.6897 & 0.3782 & 0.2949 & 0.1896 \\
\bottomrule
\end{tabular}}
\end{table*}
\subsection{Detailed Breakdown by Object Type}
\label{sec:breakdown_type}

\paragraph{ModeSeq (Motion Prediction).}
Table~\ref{tab:modeseq_avg_only} summarizes the breakdown of ModeSeq across object types. Vehicles achieve the strongest ranking scores (Soft mAP v2/mAP v2 = 0.5181/0.5095) and the lowest overlap rate (0.0465). Pedestrians obtain the lowest displacement errors (Min ADE/Min FDE = 0.3407/0.7117) and the lowest miss rate (0.0776), while exhibiting the highest overlap rate (0.2617). Cyclists remain the most challenging category, with the highest miss rate (0.1706) and larger errors (Min ADE/Min FDE = 0.6852/1.4136). The aggregated \textit{All} row reports overall performance (Soft mAP v2/mAP v2 = 0.4737/0.4665, Min ADE/Min FDE = 0.5680/1.1766, Miss Rate = 0.1204, Overlap Rate = 0.1275).

\paragraph{Parallel ModeSeq (Interaction Prediction).}
Table~\ref{tab:parallel_modeseq_avg_only} reports the corresponding breakdown for Parallel ModeSeq. Vehicles again achieve the strongest ranking performance (Soft mAP/mAP/mAP v2 = 0.4107/0.4063/0.4063) and the lowest overlap rate (0.1328). Pedestrians yield the lowest errors (Min ADE/Min FDE = 0.6093/1.3241) but have a higher miss rate (0.3598) and the highest overlap rate (0.2909). Cyclists are the hardest category with the lowest ranking scores (Soft mAP/mAP/mAP v2 = 0.1768/0.1754/0.1754) and the highest miss rate (0.4904), together with the largest errors (Min ADE/Min FDE = 0.9164/2.0585). The \textit{All} row summarizes overall results (Soft mAP/mAP/mAP v2 = 0.2978/0.2949/0.2949, Min ADE/Min FDE = 0.7707/1.6897, Miss Rate = 0.3782, Overlap Rate = 0.1896).

\section{Conclusion}
We presented Mode-as-Sequence, a unified framework that models multimodal motion forecasting by turning an unordered mode set into an ordered mode sequence, enabling explicit mode-to-mode dependency modeling under sparse supervision. Within this framework, we developed ModeSeq as a sequential decoder that generates diverse, well-ranked hypotheses by conditioning each mode on previously generated ones, and Parallel ModeSeq as an efficient parallel realization that preserves the same causal dependency through masked mode-to-mode attention. To train reliable hypothesis sets and confidence scores from single ground-truth futures, we introduced EMTA and its joint-scene extension MA-EMTA, further strengthened by a lightweight ranking regularizer that reduces confidence inversions. Extensive experiments and challenge submissions demonstrate that the proposed decoding and training principles consistently improve ranking quality, coverage, and accuracy across datasets and horizons, while providing practical scalability via parallel decoding and mode extrapolation. In the Waymo Open Dataset challenges, our submissions achieved first-place results (ModeSeq won the 2024 LiDAR-free Motion Prediction Challenge, and Parallel ModeSeq won the 2025 Interaction Prediction Challenge), highlighting the effectiveness of Mode-as-Sequence for both accurate sparse multimodal prediction and efficient large-scale deployment.
\bibliographystyle{IEEEtran}
\bibliography{main}

\newpage

 




\vfill

\end{document}